\documentclass[12pt,fleqn,a4paper]{article}

\usepackage{authblk}
\usepackage[twoside,bindingoffset=2.1cm,includeheadfoot,textwidth=17cm,textheight=26.5cm,top=0.5cm]{geometry}
\usepackage{amssymb,array,delarray,verbatim,alltt,epsfig,float,amsthm}
\usepackage{graphicx,algorithmicx,algorithm,algpseudocode,xspace,xcolor}
\usepackage[centertags]{amsmath}
\usepackage{wrapfig}
\usepackage{amsfonts,fancyvrb,cancel}
\usepackage{subfigure,dsfont}
\usepackage{tikz}
\usetikzlibrary{arrows,shapes,backgrounds,through,shadows,patterns,mindmap}
\usetikzlibrary{calc}
\graphicspath{{./matlab/}{./}{./figures/}{./../figures/}{./../matlab/}{./../matlab/figures/}{./../codeOO/misc/}}

\newcommand{\csett}[1]{\sett{#1}^c}

\newcommand{\beq}{\begin{equation}}
\newcommand{\eeq}{\end{equation}}

\newcommand{\cb}[1]{\left\{ {#1} \right\}}
\newcommand{\br}[1]{\left( {#1} \right)}
\newcommand{\sq}[1]{\left[ {#1} \right]}

\renewcommand{\eqref}[1]{equation(\ref{#1})}

\newcommand{\figref}[1]{fig(\ref{#1})}

\newcommand{\secref}[1]{section(\ref{#1})}

\newcommand{\ave}[2]{\mathbb{E}_{#2}\br{#1}}

\newcommand{\sgn}[1]{{\rm sgn}\br{#1}}

\newcommand{\bigo}[1]{O\br{#1}}

\newcommand{\ie}{{\em i.e. \xspace}}

\newcommand{\minus}{\backslash}

\newcommand{\myset}[1]{\mathcal{\uppercase{#1}}}

\newcommand{\sett}[1]{\myset{#1}}

\newcommand{\ocm}{\hspace{1cm}}
\newcommand{\hcm}{\hspace{0.25cm}}

\renewcommand{\v}[1]{\mathbf{#1}}

\newcommand{\trans}{^{\textsf{T}}}

\newcommand{\tw}{\textwidth}
\newcommand{\newl}{\hspace{1mm}\\}

\newcommand{\ind}[1]{\mathbb{I}\sq{#1}}

\title{Dealing with a large number of classes -- Likelihood, Discrimination or Ranking? }
\author{David Barber, Aleksandar Botev}
\affil{Department of Computer Science\\
University College London}
\date{\today}

\begin{document}

\maketitle
\begin{abstract}
We consider training probabilistic classifiers in the case of a large number of classes. The number of classes is assumed too large to perform exact normalisation over all classes. To account for this we consider a simple approach that directly approximates the likelihood. We show that this simple approach works well on toy problems and is competitive with recently introduced alternative non-likelihood based approximations. Furthermore, we relate this approach to a simple ranking objective. This leads us to suggest a specific setting for the optimal threshold in the ranking objective.
\end{abstract}

%LETS RUN AN EXPERIMENT TO TEST WHETHER THE GRADIENTS FOR CLASSES IN THE MINIBATCH IS MORE ACCURATE THAN FOR CLASSES OUTSIDE THE MINIBATCH.

\section{Probabilistic Classifier}

Given an input $x$, we define a distribution over class labels $c\in\cb{1,\ldots,C}$ as
\beq
p_\theta(c|x) = \frac{u_\theta(c,x)}{Z_\theta(x)}, \ocm u_\theta\geq 0
\eeq
with normalisation
\beq
Z_\theta(x)\equiv \sum_{d=1}^C u_\theta(d,x)
\eeq
Here $\theta$ represents the parameters of the model. A well known example is the softmax model in which $u_\theta(c,x)=\exp\br{s_\theta(c,x)}$, with a typical setting for the score function $s_\theta(c,x)=\v{w}_c\trans\v{x}$ for input vector $\v{x}$ and parameters $\theta=\cb{\v{w}_1,\ldots,\v{w}_C}$. The normalisation requires summing over all classes $c$ and the assumption is that this will be prohibitively expensive.\newl 

Computing the exact probability requires the normalisation to be computed over all $C$ classes and we are interested in the situation in which the number of classes is large. For example, in language models, it is not unusual to have of the order of $C=100,000$ classes, each class corresponding to a specific word.  This causes a bottleneck in the computation.  In language modelling several attempts have been considered to alleviate this difficulty. Early attempts were to approximate the normalisation by Importance sampling  \cite{Bengio2003,Bengio2008}. Alternative, non-likelihood based training approaches such as Noise Contrastive Estimation \cite{Gutmann2010,Mnih2012}, Negative Sampling \cite{mikolov2013} and BlackOut \cite{Blackout} have been considered. Our conclusion is that none of these alternatives is superior to very simple likelihood based approximation.

\subsection{Maximum Likelihood}

Given a collection of data $\sett{D}\equiv\cb{(x_n,c_n),n=1,\ldots,N}$ a natural\footnote{Maximum Likelihood has the well-known property that is is asymptotically  efficient.} way to train the model is to maximise the log likelihood $L(\theta) \equiv \sum_{n=1}^N  L_n(\theta)$ where the log likelihood of an individual datapoint is
\beq
L_n(\theta) = \log p_\theta(c_n|x_n) = \log u_\theta(c_n,x_n)-\log Z_\theta(x_n)
\eeq
The gradient $g(\theta)\equiv \partial_\theta L(\theta)$ is given by $g(\theta) = \sum_{n=1}^N g_n(\theta)$ where the gradient associated with an individual datapoint $n$ is given by
\beq
g_n(\theta)={\partial_\theta\log u_\theta(c_n,x_n)-\frac{1}{Z_\theta(x_n)}\sum_{d=1}^C \partial_\theta u_\theta(d,x_n)}
\eeq
We can write this as
\beq
g_n(\theta)= \sum_{c=1}^C \br{\delta(c,c_n)-p_\theta(c|x_n)}\partial_\theta \log u_\theta(c,x_n)
\label{eq:g:batch}
\eeq
where the Kronecker delta $\delta(c,c_n)$ is 1 if $c=c_n$ and zero otherwise. This has the natural property that when the model $p_\theta(c|x_n)$ predicts the correct class label 
$c_n$ for each datapoint $n$, then the gradient is zero.  The gradient is a weighted combination of gradient vectors, 
\beq
g_n(\theta)= \sum_{c=1}^C \gamma_n(c)\partial_\theta \log u_\theta(c,x_n)
\label{eq:grad}
\eeq
with weights given by
\beq
\gamma_n(c)\equiv \delta(c,c_n)-p_\theta(c|x_n)
\eeq
(dropping the notational dependence of $\gamma$ on $\theta$ for convenience). Since $p\in\sq{0,1}$, we note that $\gamma_n(c)\in\sq{-1,1}$.\newl 

In practice, rather than calculating the gradient on the full batch of data, we use a much smaller randomly sampled minibatch $\sett{M}$ of datapoints, $|\sett{M}|\ll N$ (typically of the order of 100 examples) and use the minibatch gradient
\beq
g(\theta)= \sum_{m\in\sett{M}}g_m(\theta)
\eeq
to update the parameters at each iteration (for example by Stochastic Gradient Ascent).

\subsection{Class specific parameter models}

We note that in the special case of a `class specific model' $u_\theta(c,x_n)=\tilde{u}_{\theta_c}(x_n,\phi)$ for class specific parameters $\theta_c$ and class independent parameters $\phi$, the gradient contribution wrt $\theta_a$, is
\beq
g_n(\theta_a)= \sum_{c=1}^C \gamma_n(c)\partial_{\theta_a} \log \tilde{u}_{\theta_c}(x_n,\phi)
= \gamma_n(a)\partial_{\theta_a} \log \tilde{u}_{\theta_a}(x_n,\phi)
\label{eq:grad:a}
\eeq
so that at least one troublesome summation over all classes is removed; note however that $\gamma$ still requires a summation over all classes. 
%Hence the sum over all classes is replaced by the sum over the union of the classes contained in the minibatch $\sett{C}\equiv \cup \cb{c_m}, m\in\sett{M}$. All other gradients wrt $\theta_a$ (for $a$ not in the minibatch class) are zero. This gives
%\beq
%g(\theta)= \sum_{m\in\sett{M}}\sum_{c\in\sett{C}} \gamma_m(c)\partial_\theta \log u_{\theta_c}(x_m)
%\eeq
%%In this case, the problematic summation over all classes in \eqref{eq:g:batch} is removed. 
%%For models which are not of this form, we suggest an additional sampling approximation in \appref{sec:general:case}.
A classic example of a class specific model is a network classifier with a softmax output layer 
\beq
p(c|\v{x}) = \frac{e^{\v{w}_c\trans \v{h}_\phi(\v{x})}}{\sum_{d=1}^C e^{\v{w}_d\trans \v{h}_\phi(\v{x})}}
\label{eq:class:spec:gradient}
\eeq
where $\v{h}_\phi(\v{x})$ is the value of the final hidden layer of the network. 
%This gives
%\beq
%u(c,\v{x})=e^{\v{w}_c\trans \v{h}_\phi(\v{x})}
%\eeq 
In general, even though a minibatch $\sett{M}$ may only contain labels for a small set of classes, the log-likelihood contribution depends (through $\gamma$) on all $C$ classes.  Whilst, perhaps ideally, one would try to directly approximate the gradient \eqref{eq:class:spec:gradient} using a small number of classes, we would need to decide which of \emph{all} classes $1,\ldots,C$ we would wish to update.  
In terms of monitoring the likelihood it is also useful to have an approximation of the likelihood itself, dependent on only a small number of the total classes.
%For this reason, it is natural to consider on approximations to the log-likelihood that themselves only depend on a small number of classes, since in this case the only a small number of class parameters need to be updated for each minibatch.

\section{Normalisation Approximation}
For a likelihood approximation we require an approximation of the normalisation
\beq
Z_\theta(x_m)= \sum_{d=1}^C u_\theta(d,x_m)
\eeq
A common way to address this problem is by Importance sampling  \cite{Geyer94,Bengio2003}, which is based on the approximation
\beq
Z_\theta(x_m) = \sum_{d=1}^C q(d)\frac{u_\theta(d,x_m)}{q(d)} =\ave{\frac{u_\theta(d,x_m)}{q(d)}}{q} 
\eeq
for some Importance distribution $q(d)$.  By drawing $S$ samples $d_1,\ldots,d_S$, $d_s\in\cb{1,\ldots,C}$ from $q$, a sampling approximation is then given by
\beq
Z_\theta(x_m) \approx \frac{1}{S}\sum_{s=1}^S \frac{u_\theta(d_s,x_m)}{q(d_s)} 
\eeq
In practice, however, this approach is problematic since the variance of this estimator is high \cite{Bengio2008}. In particular, the approximated scalar
\beq
\gamma_m(c)\approx \delta(c,c_m) - \frac{u_\theta(c,x_m)}{\frac{1}{S}\sum_{s=1}^S \frac{u_\theta(d_s,x_m)}{q(d_s)}}
\eeq
is no longer bounded between $-1$ and $1$. This can create highly varying and inaccurate gradient updates -- indeed, the gradient direction $g_m(\theta)$ is not guaranteed to be correct, neither for the general model nor the class specific model. Whilst there have been attempts to adapt $q$ to reduce the variance in $\gamma$, these are typically significantly more complex and may require signficant computation to correct wild gradient esimates     \cite{Bengio2008}. 
%In particular, the method in \cite{Bengio2008} is specific to sequential prediction, and inapplicable in the general case. 
We seek therefore an alternative approach for this general class of models.

\subsection{Sampling approximations\label{sec:our:approach}}

For each datapoint $n$ we define a set of classes $\sett{C}_n$ that must be explicitly summed over in forming the approximation. This defines then for each datapoint $n$ a complementary set of classes  $\csett{C}_n$, (all classes except for those in $\sett{C}_n$). We can then write
\begin{align}
Z_\theta(x_n) &= \sum_{c\in\sett{C}_n} u_\theta(c,x_n)+\sum_{d\in\csett{C}_n} u_\theta(d,x_n)
%&=\sum_{c\in\sett{C}} u_\theta(c,x_n)+\kappa\ave{\sum_{d\in\csett{C}} s_du_\theta(d,x_n)}{\v{s}\tilde p}
\end{align}
We propose to simply approximate the sum over the complementary classes by sampling.  However, in order to ensure that this results in an approximate $-1\leq\gamma_n(c)\leq 1$, we require that $\sett{C}_n$ contains the correct class $c_n$. This simple setting therefore significantly reduces the variance in the sampling estimate of the gradient.
Surprisingly, whilst there have been several closely related suggestions, we are not aware  of any previous approaches taking this route. We will show that this approach leads to a simple and effective way to approximate the gradient, and also suggests connections to ranking based approaches.

\subsubsection{Importance Sampling}
We consider the general problem of summing over a collection of elements 
\beq
Z = \sum_{i=1}^C z_i
\eeq
An obvious approximation is to use Importance sampling from a distribution $q(i)$. Based on the identity $Z=\sum_i q(i) z_i/q_i$ we draw $S$ samples from $q$ and form the approximation
\beq
\tilde{Z} = \frac{1}{S}\sum_{s=1}^S \frac{z_s}{q(s)}
\label{eq:Ztilde:IS}
\eeq
Whilst an unbiased estimator of $Z$, the variance of the Importance sampler is
\beq
\frac{1}{S}\br{\sum_{c=1}^C\frac{z_c^2}{q(c)} - Z^2}
\eeq
A downside of Importance sampling is that, even as the number of samples $S$ is increased beyond the number of elements $C$ in the exact sum, $\tilde{Z}$ remains an approximation, despite the approximation method using more computation than the exact calculation would require.  For this reason, we also consider a sampling approach with bounded computation. 

\subsubsection{Bernoulli Sampling}

An alternative to Importance sampling is to consider the identity
\beq
Z = \sum_{i=1}^C z_i = \ave{\sum_{i=1}^C \frac{s_i}{b_i}z_i}{\v{s}\sim \v{b}}
\eeq
where each independent Bernoulli variable $s_i\in\cb{0,1}$ and $p(s_i=1)=b_i$. Unlike Importance sampling, no samples can be repeated. We propose to take a single joint sample from $\v{s}$ to form the Bernoulli sample approximation
\beq
Z \approx  \sum_{i=1: s_i=1}^C \frac{z_i}{b_i}
\label{eq:Ztilde:Bernoulli}
\eeq
This Bernoulli sampler of $Z$ is unbiased with variance\footnote{For the same computational cost this variance will typically be lower than the variance of the IS. To illustrate this, consider that the $z_i$ are drawn from a distribution with mean $\mu$ and variance $\sigma^2$, and that the setting of the Bernoulli sampling probabilities $b_i$ and the IS weights $q$ do not depend on the value of $z$. Then the IS has expected variance
\[
\frac{1}{S}\br{\br{\sigma^2+\mu^2}\sum_c \frac{1}{q(c)}-\mu^2}
\]
and the Bernoulli sampler has expected variance
\[
\br{\sigma^2+\mu^2}\sum_c\br{\frac{1}{b_c}-1}
\]
In this setting, the minimal variance for IS is given by $q(c)=1/C$. To ensure that both the IS and Bernoulli sampler use a similar amount of computation, for the Bernoulli sampling we set $b$ such that the expected number of samples is $S$, giving $b_c=S/C$. If we assume also that the number of samples is only a fraction $f$ of the total number of classes, $S=fC$ for $0\leq f\leq 1$ then the IS has variance
\[
\br{\sigma^2+\mu^2}\frac{C}{f}-\frac{\mu^2}{fC}
\]
whilst the Bernoulli sampler has variance
\[
\br{\sigma^2+\mu^2}\frac{C}{f}-\br{\sigma^2+\mu^2}C
\]
For small $f$ and large $C$, the variance from the Bernoulli sampler can therefore be significantly lower than for the IS. 
}
\beq
\sum_{c=1}^C \br{\frac{1}{b_c}-1}z_c^2
\eeq
For $b_i\rightarrow 1$, each $s_i$ is sampled in state 1 with probability 1 and the approximation recovers the exact summation. In our experiments we set each $b_i$ such that the expected number of samples from the complementary set is equal to a user defined value $K$. Specifically we set $b_c = f(c)^\alpha$ where $f(c)$ is the empirical frequency of observed classes in the whole training set and choose $\alpha$ such that $\sum_{d\in\csett{C}_n} b_c=K$. For $K=|\csett{C}_n|$, this gives $\alpha=0$ and all classes are summed over, giving the exact result. For $K<|\csett{C}_n|$ then $\alpha<1$ and we sum over a subset of all complementary classes, including those classes that occur more frequently in the training set with higher probability.  The variance of the number of samples used in the Bernoulli sampler is $\sum_{c\in\csett{C}_n} b_c\br{1-b_c}\leq K$. 

\subsubsection{The approximate gradient}

Both Importance and Bernoulli sampling therefore give approximations in the form
%We consider any distribution $p(\v{s})$ over a set of $|\csett{C}|$ variables %$s_d\in\cb{0,1}$ with uniform marginal probability $p(s_d=1)$.  Then 
%\begin{align}
%Z_\theta(x_n) &=\sum_{c\in\sett{C}_n} u_\theta(c,x_n)+\ave{\sum_{d\in\csett{C}_n} %\frac{s_d}{\ave{s_d}{}}u_\theta(d,x_n)}{\v{s}\sim p}
%\end{align}
%From this we draw a single sample $\v{s}$ from $p(\v{s})$ which gives a `negative' set of indices $\sett{N}$. This forms our approximation
\beq
\tilde{Z}_\theta(x_n) = \sum_{c\in\sett{C}_n} u_\theta(c,x_n)+\sum_{d\in\sett{N}_n}{\kappa_{d,n}u_\theta(d,x_n)}
\eeq
where $\sett{N}_n$ is a set of $K$ `negative' sampled classes from the complementary set. In the Importance case, $\kappa_{d,n}=1/(Kq_n(d))$ represents the probability of sampling class $d$ according to the IS distribution\footnote{The IS distribution $q_n(d)$ depends on the data index $n$, since the IS distribution must not include the classes in the set $\sett{C}_n$.} $q_n(d)$; in the Bernoulli case $\kappa_{d,n}=1/b_{d,n}$ is the probability that $s_d=1$. 
%In our experiments we sample a fixed number $\nu$ of negative samples from the full set of $\csett{C}$ negative samples which corresponds to using the weight
%\beq
%\kappa=\frac{|\csett{C}|}{\nu}
%\eeq
This gives an approximate log likelihood contribution
\beq
\tilde{L}_n(\theta)=u_\theta(c_n,x_n) - \log \tilde{Z}_\theta(x_n)
\eeq
with derivative
%\beq
%\partial_\theta \tilde{L}_n(\theta)=\partial_\theta u_\theta(c_n,x_n) - \frac{1}{\tilde{Z}_\theta(x_n)}\br{\sum_{c\in\sett{C}} \partial_\theta u_\theta(c,x_n)+\kappa\sum_{d: s_d=1}\partial_\theta u_\theta(d,x_m)}
%\eeq
%
%\beq
%\partial_\theta \tilde{L}_n(\theta)=\partial_\theta u_\theta(c_n,x_n) - \sum_{c\in\sett{C}} \tilde{p}(c|x_n)\partial_\theta \log u_\theta(c,x_n)-\sum_{d: s_d=1}\tilde{p}(d|x_n)\partial_\theta \log u_\theta(d,x_n)
%\eeq
%where $\sett{N}$ are the negative classes for the datapoint (or minibatch?). 
\beq
\partial_\theta \tilde{L}_n(\theta)=\sum_{c\in\sett{C}_n'}\br{\delta(c,c_n)- \tilde{p}(c|x_n)}\partial_\theta \log u_\theta(c,x_n)
\eeq
where
\beq
\tilde{p}(c|x_n)= \left\{\begin{array}{ll}
u_\theta(c,x_n)/\tilde{Z}_\theta(x_n) & c\in\sett{C}_n\\
\kappa_{c,n} u_\theta(c,x_n)/\tilde{Z}_\theta(x_n) & c\in\sett{N}_n\\
\end{array}\right.
\label{eq:tilde:p}
\eeq
Note that $\tilde{p}$ is a distribution\footnote{In \cite{Jean2015} Importance sampling is used to motivate an approximation that results in a distribution over a predefined subset of the classes. However the approximation is based on a biased estimator of the normalisation and as such is not an Importance sampler in the standard sense.} over the classes $\sett{C}_n'=\sett{C}_n\cup\sett{N}_n$. 
Hence the approximation 
\beq
\tilde{\gamma}_m(c) \equiv \delta(c_m,c)-\tilde{p}(c|x_m) 
\eeq
has the property $\tilde{\gamma}_m(c)\in\sq{-1,1}$. \newl 
%
%For a softmax layer model, this depends only on the weights $\v{w}_c$ that are present in the minibatch and negative samples, so that we would only need to update these parameters (in addition to the other network parameters). This is the same in the NCE approach. \newl
%
%An alternative is to form the approximation
%\beq
%\tilde{Z}_\theta(x_n) = u_\theta(c_n,x_n)+\kappa \sum_{d\in\sett{N}(c_n)}u_\theta(d,x_n)
%\eeq
%where $\sett{N}(c_n)$ is a sampled set of all classes $1,\ldots,C$ except for $c_n$, giving $\kappa=|\csett{C}|/|\sett{N}(c_n)|$. 
%%An alternative to the approximation for $\tilde{p}$ is to write
%%\beq
%%{p}(c|x_m) = \frac{u_\theta(c,x_m)}{u_\theta(c,x_m)+\sum_{d\neq c}u_\theta(d,x_m)}
%%\label{eq:p:exact:c}
%%\eeq
%%and similarly approximate $\sum_{d\neq c}u_\theta(d,x_m)$ by Bernoulli sampling.  
%This also gives a corresponding approximate distribution $\tilde{p}(c|x_m)$ that is bounded between 0 and 1.\newl  
%%Depending on $\kappa$, this will typically however result in a higher variance (but lower computational cost) than \eqref{eq:tilde:p} since the variance of the residual sum over the negative classes is greater. For this reason we restrict attention in our experiments to the approximation \eqref{eq:tilde:p}.\newl

For fixed $\kappa$, the bias\footnote{The estimator of $p(c|x_n)$ is biased since the estimator of the inverse normalisation $1/Z_\theta$ is biased. One can form an effectively unbiased estimator of $1/Z_\theta$ by a suitable truncated Taylor expansion of $1/Z$, see \cite{booth2007}. However, each term in the expansion requires a separate independent joint sample from $p(\v{s})$ (for the BS) and as such is sampling intensive, reducing the effectiveness of the approach.} in estimating $p(c|x_m)$ (which depends on $1/Z$) is given by the Taylor expansion
\beq
\tilde{p}(c|x_m)=p(c|x_m)\br{1+(1-\kappa)p(\sett{N}_m|x_m)}+\bigo{(1-\kappa)^2}
\eeq
where $p(\sett{N}_m|x_m)=\sum_{d\in\sett{N}_m}p(d|x_m)$ is the total probability of the sampled negative classes. For a well trained model, the probability of generating incorrect labels for the minibatch will be low, resulting in a low bias for the gradient.
In the limit of a large number of Importance samples, $\kappa\rightarrow 1$; similarly for Bernoulli sampling, as $K$ tends to the size of the complementary set, $\kappa\rightarrow 1$; the resulting estimators of $p(c|x_n)$ are therefore consistent.\newl

%For IS this follows from the standard argument that the variance of the estimator of $Z$ and hence since for a single joint sample $\v{s}$, in the limit $b_c\rightarrow 1$ all complementary classes are included in the summation, giving an exact evaluation. \newl

For class specific parameter models, the property $\tilde{p}\in\sq{0,1}$ not only ensures a low-variance estimator of the gradient, but also results in the pleasing property that for parameters in the minibatch class ($\theta_a$ for $a\in\sett{C}$)  the sign of the gradient for each datapoint in the minibatch is correct. This follows immediately from the observation 
\beq
\sgn{\tilde{\gamma}_m(a)}=\sgn{\gamma_m(a)}
\eeq
Whilst this does not guarantee that the sign of the overall gradient approximation
\beq
\tilde{g}(\theta_a)= \sum_{m\in\sett{M}} \tilde{\gamma}_m(a)\partial_{\theta_a} \log u_{\theta_a}(x_m)
\eeq
is correct, in practice we find that the sign of the approximate minibatch gradient $\sgn{\sq{\tilde{g}}_i}$ is correct for the majority of the components of the vector. \newl 

In implementing the sampling approach, there remains the choice of the set $\sett{C}_n$. For a budget of $T$ classes to be used one could use for example $T-K$ classes for the explicit sum over $\sett{C}_n$ with the remaining $K$ classes sampled from $\csett{C}_n$. The optimal choice between using explicit sums and sampled classes to approximate $Z$ will inevitably be problem and implementation dependent. The closest comparator to BlackOut is to use a single class $\sett{C}_n=\cb{c_n}$ and sample the remainder from $\csett{C}_n$. In this case, every member of the minibatch has a corresponding set of $K$ additional samples. Depending on the details of the implementation, accessing roughly $|\sett{M}|K$ class parameters may be too expensive. For this reason, alternatives such as choosing a fixed set of classes in advance may improve efficiency of memory access. Similarly, there is a choice as to which class parameters to update for each minibatch. For example, one may update only the parameters of the observed classes in the minibatch, or all classes from the minibatch and sampled negative classes. Again, the optimal setting will be problem and implementation dependent.

\section{Relation to other approaches}

The closest approach to ours are those taken by \cite{Bengio2003,Bengio2008} which use the maximum likelihood objective, approximated by Importance sampling. Since this has previously been perceived to be impractical, alternative approaches have been considered that are either not based on maximising an approximated log likelihood or maximising the likelihood of a different model.  

%\subsection{Introduction: Softmax Models}
%
%A softmax classifier has probability 
%\beq
%p(c|\v{x}) = \frac{e^{\v{w}_c\trans\v{x}}}{\sum_{d}e^{\v{w}_{d}\trans\v{x}}}
%\eeq
%Then the log probability is
%\beq
%L\equiv\log p(c|\v{x}) = \v{w}_c\trans\v{x} - \log\sum_{d}e^{\v{w}_{d}\trans\v{x}}
%\eeq
%with gradient
%\beq
%\frac{\partial \log p(c|\v{x})}{\partial \v{w}_a} = \br{\delta_{c,a}-\frac{e^{\v{w}_a\trans\v{x}}}{\sum_{d}e^{\v{w}_{d}\trans\v{x}}}}\v{x}
%\label{eq:grad}
%\eeq
%More generally, for a set of datapoints $\br{\v{x}_n,c_n}$, $n=1,\ldots, N$, the log likelihood gradient (assuming iid data) is
%\beq
%\frac{\partial \sett{L}}{\partial \v{w}_a} = \sum_{n=1}^N \br{\delta_{c_n,a}-\frac{e^{\v{w}_a\trans\v{x}_n}}{\sum_{d}e^{\v{w}_{d}\trans\v{x}_n}}}\v{x}_n
%\eeq
%In practice, we compute the gradient for a randomly selected minibatch of $M<N$ datapoints. 

%In the case of a large number of classes, the sum $\sum_{d}e^{\v{w}_{d}\trans\v{x}_n}$ is expensive. There are a variety of approaches that can be used here. 

\subsection{Hierarchical Softmax}

Hierarchical softmax \cite{MorinBengio2005} defines a binary tree such that the probability of a leaf class is the product of edges from the root to the leaf. For the softmax regression setting $u(c,\v{x}_m)=\exp\br{\v{w}_c\trans\v{x}_m}$, each (left) edge child of a node $n$ is associated with a probability $\sigma\br{\v{w}_n\trans\v{x}}$, with a corresponding weight $\v{w}_n$ for each node. The advantage of this is that it defines a distribution over all $C$ classes and removes the requirement to explicitly normalise over all classes.  The probability of an observed class then scales with $\log C$, rather than $C$ in the standard softmax approach. A disadvantage is that, apart from the additional implementation complexity, the number of parameters is significantly larger than in the standard softmax, with one parameter per node in the tree. Whilst this can be addressed by parameter sharing, hierarchical softmax defines a new model, rather than an approximation to the original softmax model. As such we will not consider it further here.
%
%Below we consider a simple method to find a fast approximate gradient computation of the standard softmax. It is useful to write
%\beq
%\frac{\partial \sett{L}}{\partial \v{w}_a} = \sum_{n=1}^N \gamma_{n,a}\v{x}_n
%\eeq
%where
%\beq
%\gamma_{n,a} \equiv \delta_{c_n,a}-\frac{e^{\v{w}_a\trans\v{x}_n}}{\sum_{d}e^{\v{w}_{d}\trans\v{x}_n}} \in\sq{-1,1}
%\eeq
%We note that $\gamma_{n,a}$ is positive when $c_n=a$ and is negative otherwise.\newl
%
%When the data vectors $\v{x}_n$, $n=1\ldots,N$ are linearly independent, the gradient is zero when each $\gamma_{n,a}$ is zero. This would require that the length of the vectors $\v{w}_a$ becomes unbounded.  
%
%
%
%\section{A normalisation approximation}
%
%
%A simple approximation is to consider the unique classes present in a minibatch of data, and append these with a small set of additionally randomly selected classes from $1,\ldots, C$. This forms a set of classes $\sett{C'}$ (which therefore includes the classes in the minibatch and some additional randomly selected classes). We will update the parameters for the classes $\sett{C'}$ using the approximation
%\beq
%\tilde{\gamma}_{m,a} \equiv \delta_{c_m,a}-\frac{e^{\v{w}_a\trans\v{x}_m}}{\sum_{d\in\sett{C'}}e^{\v{w}_{d}\trans\v{x}_m}} 
%\eeq
%where $m$ indexes the datapoint in the minibatch. Since we update only the parameters in $\sett{C'}$, then $a\in\sett{C'}$ and $\tilde{\gamma}_{m,a}\in\sq{-1,1}$. 
% 

\subsection{Noise Contrastive Estimation}

%Need to implement this. It's a bit similar to negative sampling. Odd thing is that it needs an infinite number of noise samples to tend to the max likelihood gradient (whereas the normalisation approach has bounded complexity to get the exact likelihood). 

NCE \cite{Gutmann2010,Gutmann2012} is a general approach that can be used to perform estimation in unnormalised probability models and has been successfully applied in the context of language modelling in \cite{Mnih2012,Mnih2013}. The method generates data from the `noise' classes (which range over all classes, not just the negative classes) for each datapoint in the minibatch. The objective is related to a supervised learning problem to distinguish whether a datapoint is drawn from the data or noise distribution. The method forms a consistent estimator of $\theta$ in the limit of an infinite number of samples from a noise distribution $p(c)$ $c\in\cb{1,\ldots,C}$. \newl

The method has gradient for minibatch datapoint $(c_m,x_m)$
\beq
\frac{kp_n(c_m)}{p'_\theta(c_m|x_m)+kp_n(c_m)}\partial_\theta\log p'_\theta(c_m|x_m)-\sum_{i=1}^k \frac{p'_\theta(d_i)}{p'_\theta(d_i|x_m)+kp_n(d_i)}\partial_\theta\log p'_\theta(d_i|x_m)
\eeq
where
\beq
p'_\theta(c_m|x_m)=u_\theta(c_m,x_m)/z_m
\eeq
and the total gradient sums the gradients over the minibatch. 
The method requires that each datapoint in the minibatch has a corresponding scalar parameter $z_m$ (part of the full parameter set $\theta$) which approximates the normalisation $Z_m(\theta)$. Formally, the objective is optimised when $z_m=Z_m(\theta)$ which would require an expensive inner optimisation loop for each minibatch over these parameters. For this reason, in practice, these normalisation parameters are set to $z_m=1$ \cite{Mnih2012}. Formally speaking this invalidates the consistency of the approach unless the model is rich enough that it can implicitly approximate the normalisation constant\footnote{This is the assumption in \cite{Mnih2012} in which the model is assumed to be powerful enough to be `self normalising'.}. In the limit of the number of noise samples tending to infinity, the optimum of the NCE objective coincides with maximum likelihood optimum. A disadvantage of this approach therefore compared to Bernoulli sampling is that (in addition to the formal requirement of optimising over the $z_m$) the method requires in principle an infinite amount of computation to match the maximum likelihood objective.  Whilst this method has been shown to be effective for complex `self normalising' models, in our experiments with softmax regression, this approach (setting $z_m=1$) performs very poorly and does not lead to a practically usable algorithm.

%
%\subsection{A special case and relation to NCE}
%
%Consider the approximation
%\beq
%\frac{e^{\v{w}_a\trans\v{x}_n}}{\sum_d e^{\v{w}_a\trans\v{x}_n}}
%=\frac{e^{\v{w}_a\trans\v{x}_n}}{e^{\v{w}_a\trans\v{x}_n} + \sum_{d\neq a} e^{\v{w}_a\trans\v{x}_n}}
%=\frac{e^{\v{w}_a\trans\v{x}_n}/Z}{e^{\v{w}_a\trans\v{x}_n}/Z + \sum_{d\neq a} e^{\v{w}_a\trans\v{x}_n}/Z}
%=\frac{p(a|x_n)}{p(a|x_n) + p(d\neq a|x_n)}
%\eeq
%where $p(a|x_n) + p(d\neq a|x_n)=1$. Note that the term $p(d\neq a|x_n)$ can be consider the probability of generating a negative label.  One can also write this term as
%\beq
%p(d\neq a|x_n) = \sum_{d\neq a} p(d|x_n)= \frac{\sum_{d\neq a} e^{\v{w}_d\trans\v{x}_n}}{\sum_{d} e^{\v{w}_d\trans\v{x}_n}}
%= \frac{\sum_{d\neq a} q(d) \frac{e^{\v{w}_d\trans\v{x}_n}}{q(d)}}{\sum_{d} q(d) \frac{e^{\v{w}_d\trans\v{x}_n}}{q(d)}}
%\eeq
%where $q(d)$ is an importance weight.
%
%An equivalent approach is to write
%\beq
%p(d\neq a|x_n) = \sum_{d\neq a} p(d|x_n)=  k \av{\sum_{d\neq a} s_d p(d|x_n)}_s 
%=k \av{p(d\neq a|s,x_n)}_s 
%\eeq
%where each $s_d\in\cb{0,1}$ and $p(s_d=1)=1/k$ for some constant $c\in(0,1)$.  The term $\av{p(d\neq a|s,x_n)}_s$ is simply the average probability of a set of noise data, in which each noise is generated with probability $1/k$. 
%
 
\subsection{Ranking approaches}

An alternative to learning the parameters of the model by maximum likelihood is to argue that, when the correct class is $c$,  we need $u_\theta(c,x)$ to be greater than $u_\theta(d,x)$ for all classes $d\neq c$. For example in the softmax regression setting $u_\theta(c,\v{x})=\exp(\v{w}_c\trans\v{x})$ we may stipulate that $\v{w}_c\trans\v{x}$ be greater than all other 
 $\v{w}_d\trans\v{x}$, for $d\neq c$, namely
\beq
\v{w}_c\trans\v{x}-\v{w}_d\trans\v{x} > \alpha
\eeq
for some positive constant $\alpha$.  This is the hinge loss ranking approach taken in  \cite{CollobertWeston2008} in which, without loss of generality, $\alpha=1$ is used.  A minor modification that results in a differentiable objective is to maximise the log ranking  
\beq
\log \sigma(\v{w}_c\trans\v{x}-\v{w}_d\trans\v{x}-\alpha)
\eeq
where $\sigma(x)=1/(1+e^{x})$ for some chosen constant $\alpha>0$. This has gradient with respect to $\v{w}_a$ given by
\beq
\br{1-\sigma\br{\v{w}_c\trans\v{x}-\v{w}_d\trans\v{x}-\alpha}}\br{\delta_{ac}-\delta_{ad}}\v{x}
\label{eq:rank:single:update}
\eeq
%Again, note that this is directionally correct for each datapoint $\v{x}$. 
For each element $m$ of a minibatch of data, we therefore use the setting
\beq
\hat{\gamma}_{m}(c) \equiv  \frac{1}{|\sett{N}_m|}\sum_{d\in\sett{N}_m}\br{1-\sigma\br{\v{w}_{c_m}\trans\v{x}_m-\v{w}_d\trans\v{x}_m-\alpha}}\br{\delta_{c,c_m}-\delta_{c,d}}
\label{eq:rank:update}
\eeq
where $\sett{N}_m$ is the set of negative classes for datapoint $m$. This encourages the overlap $\v{w}_{c_m}\trans\v{x}_m$ to be higher than $\v{w}_{d}\trans\v{x}_m$ for each negative class $d$.
% Hence, for each datapoint $m$ in the minibatch we compute a contribution to the gradient of each parameter vector $\v{w}_a$ in the minibatch. 
As before, this gives a value $\hat{\gamma}_{m,a}\in\sq{-1,1}$. It is straightforward to show that this ranking objective has a negative definite Hessian and that this corresponds therefore to a concave optimisation problem. \newl

As we will argue below, the setting $\alpha=1$ is (in general) suboptimal, and a preferable setting is $\alpha=\log(C-1)$, showing that this can make a significant difference to the bias of the estimator. 

%We note that both approximations to $\gamma$ have similar time complexities.\newl

%Alex pointed out that this may be the same as the normalisation approximation except using a single sample. Need to check this.

\subsubsection{Relation to normalisation approximation}

%If we choose only a single negative example $d$ for each minibatch member $m$, then the approximation is (neglecting the prefactor)
%\beq
%\hat{\gamma}_{m}(c) \equiv  \br{1-\sigma\br{\v{w}_{c_m}\trans\v{x}_m-\v{w}_d\trans\v{x}_m-\alpha}}\br{\delta_{c,c_m}-\delta_{c,d}}
%\label{eq:rank:update:single}
%\eeq
%
%In this special case, $u_\theta(c,x_m)=u_{\theta_c}(c,x_m)$. This means that, from \eqref{eq:grad} the derivative with respect to $\theta_a$ is given by
%\beq
%\sq{g(\theta)}_a= \sum_{m=1}^M \delta_{c_m,a}\gamma_m(a)\partial_{\theta_a} \log u_{\theta_a}(a,x_m)
%\label{eq:grad:spec}
%\eeq
A variation of our approach in \secref{sec:our:approach} is to write
\beq
p(c|x_m) = \frac{u_\theta(c,x_m)}{u_\theta(c_m,x_m)+\sum_{d\neq c_m}u_\theta(d,x_m)}
\label{eq:p:exact:c}
\eeq
and use Importance sampling with a distribution $q(d)$ over the negative classes (\ie all classes not equal to $c$) to approximate the term
\beq
\sum_{d\neq c_m} u_\theta(d,x_m)\approx \sum_d q(d)\frac{u_\theta(d,x_m)}{q(d)}
\eeq
Using a uniform distribution $q(d)=1/(C-1)$ over the $C-1$ negative classes, and drawing only a single negative sample $d_m\neq c_m$ then
\beq
\sum_{d\neq c_m} u_\theta(d,x_m)\approx (C-1) u_\theta(d_m,x_m)
\eeq
and the approximation becomes
%\begin{align}
%\tilde{p}(c|x_m) &= \frac{u_\theta(c,x_m)}{u_\theta(c,x_m)+(C-1)u_\theta(d,x_m)}
%= \frac{1}{1+(C-1)\frac{u_\theta(d,x_m)}{u_\theta(c,x_m)}}\\
%&= \frac{1}{1+\exp(\log (C-1)+\log u_\theta(d,x_m)-\log u_\theta(c,x_m)}\\
%&=\sigma\br{\log u_\theta(c,x_m)-\log u_\theta(d,x_m)-\log (C-1)}
%\end{align}
\begin{align}
\tilde{p}(c_m|x_m) &= \frac{u_\theta(c_m,x_m)}{u_\theta(c_m,x_m)+(C-1)u_\theta(d_m,x_m)}\\
&=\sigma\br{\log u_\theta(c_m,x_m)-\log u_\theta(d_m,x_m)-\log (C-1)}
\end{align}
For $\log u_\theta(c,x_m)=\v{w}_c\trans\v{x}_m$ this gives
\beq
\tilde{p}(c_m|x_m) = \sigma\br{\v{w}_{c_m}\trans\v{x}_m-\v{w}_{d_m}\trans\v{x}_m-\log(C-1)}
\eeq
with $\tilde{p}(d_m|x_m)=1-\tilde{p}(c_m|x_m)$. The gradient update then matches the ranking gradient update \eqref{eq:rank:single:update} on setting $\alpha=\log(C-1)$. One can therefore view the ranking approach as a single-sample estimate of the maximum likelihood approach. As such, we would generally expect this approach to be inferior to those given by more accurate approximations to the likelihood, such as those based on using more samples. This intuition is borne out in our experiments in \secref{sec:experiments}. \newl

\begin{figure}[t]
\begin{center}
\subfigure[]{\includegraphics[height=0.4\tw]{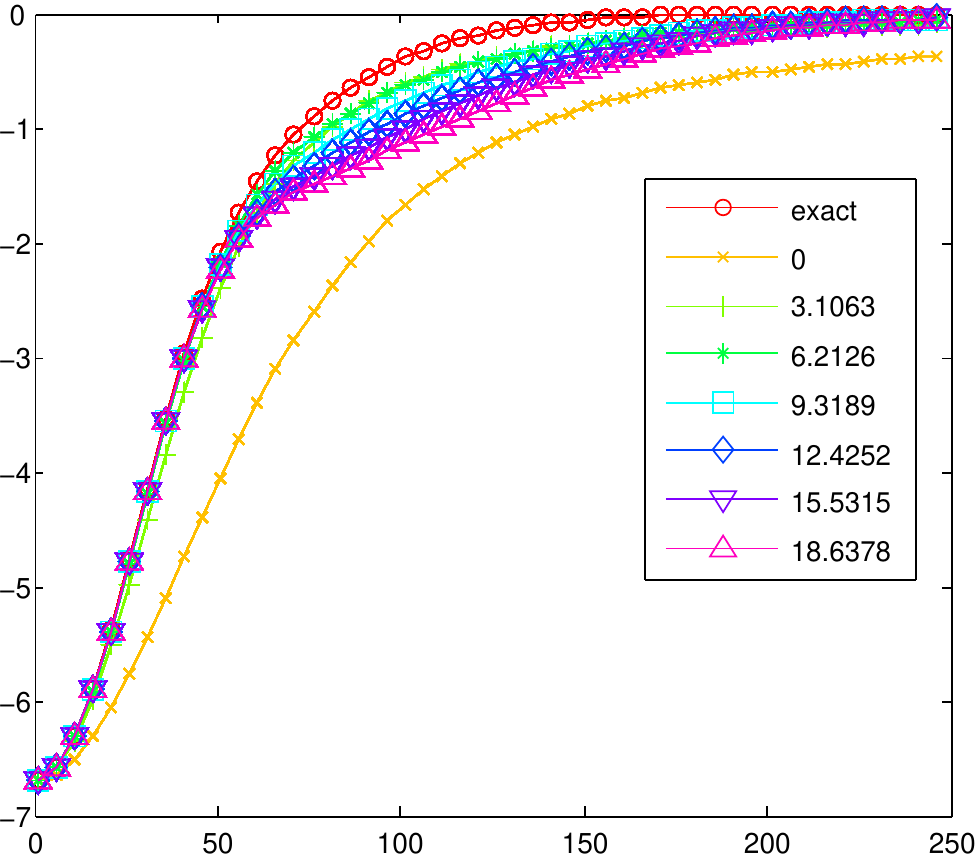}}\hcm
\subfigure[]{\includegraphics[height=0.4\tw]{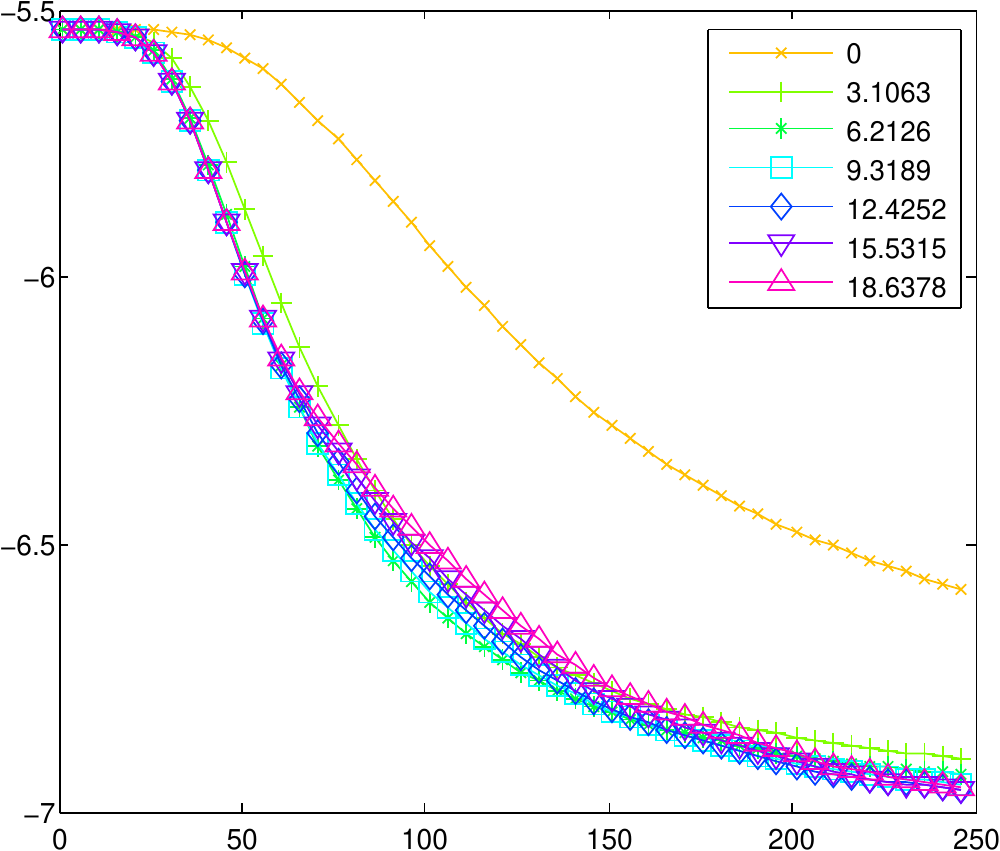}}
\end{center}
\caption{\label{fig:experiment:alphas} The experimental setting is the same as in \figref{fig:experiment}. (a) Plotted is the value of the exact log likelihood based on gradient ascent of the ranking objective for different $\alpha$ values. In this case $\log(C-1)= 6.2126$ is the suggested optimal setting. We also plot the log likelihood for the exact gradient. (b) The bias, represented as the log mean absolute difference between the class probability predictions based on the current parameters learned and the class predictions for the true underlying model (see also \secref{sec:softmax:experiments}. }
\end{figure}

%Whilst we wrote this for the case $\log u_\theta(c_m,x_m)=\v{w}_{c_m}\trans\v{x}_m$, the result holds more generally if we use the ranking criterion $\log u_\theta(c_m,x_m) -\log u_\theta(d,x_m)> \alpha$ for $d\neq c$.   
Using  the setting $\alpha=\log(C-1)$ gives the general ranking objective
\beq
\sum_{m\in\sett{M}}\sum_{d\neq c_m}\log \sigma\br{\log u_\theta(c_m,x_m) -\log u_\theta(d,x_m) - \log(C-1)}
\eeq
for subsets $d\neq c_m$ (one subset for each minibatch member) of randomly selected negative classes $d\neq c_m$. From \figref{fig:experiment:alphas} we see that the setting of $\alpha$ in the ranking objective has a strong influence on the effectiveness of the approach, with $\alpha=\log(C-1)$ being a reasonable setting, with this setting tracking the gradient more closely than other settings and giving rise to the lowest bias. Whilst the performance difference between some of the $\alpha$ settings is small, clearly the setting $\alpha=1$ is significantly less optimal than $\alpha=\log(C-1)$. 

\subsubsection{Negative Sampling}

A similar approach to ranking is to maximise $\log\sigma(\v{w}_c\trans\v{x})$ whilst minimising $\log\sigma(\v{w}_d\trans\v{x})$, for a randomly chosen subset of negative classes $d\neq c$. This is motivated in \cite{mikolov2013} as an approximation of the NCE method and has the objective
\beq
\log\sigma\br{\v{w}_c\trans\v{x}}+\sum_d \log\br{1-\sigma\br{\v{w}_d\trans\v{x}}}
\eeq
This has gradient wrt $\v{w}_a$ given by
\beq
\br{\br{1-\sigma\br{\v{w}_c\trans\v{x}}}\delta_{ac}-\sum_d\sigma\br{\v{w}_d\trans\v{x}}\delta_{ad}}\v{x}
\eeq
For ease of comparison, we scale this number of negative terms to have a similar effect to the positive term, giving for a member $m$ of the minibatch
\beq
\gamma_{m}^-(c)\equiv \frac{1}{|\sett{C}'|-1}\sum_{d\in\sett{C}'\minus c}\br{\br{1-\sigma\br{\v{w}_{c_m}\trans\v{x}_m}}\delta_{c,c_m}-\sigma\br{\v{w}_d\trans\v{x}_m}\delta_{c,d}}
\eeq
The negative sampling approach approximation to $\gamma_m(c)$ therefore has the correct sign and lies between $-1$ and $1$.  As pointed out in  \cite{mikolov2013} this objective will not, in general, have its optimum at the same point as the log likelihood. For the simple softmax regression model, we found that this approach does not yield practically useful results and as such is not considered further. The main motivation for the method is that it is a fast procedure which empirically results in useful parameters when applied in a more complex wordvec setting \cite{mikolov2013}.

\subsection{BlackOut}

The recently introduced BlackOut \cite{Blackout} is a discriminative approach based on an approximation to the true discrimination probability. This forms the approximation
\beq
\tilde{p}(c|x,\theta) = \frac{q_c u_\theta(c,x)}{q_c u_\theta(c,x) + \sum_{d\in\sett{N}}q_d u_\theta(d,x)}
\eeq
Here $q_c=1/Q(c)$ where $Q(c)$ is a distribution over all classes $1,\ldots,C$. The ratio $q_c=1/Q(c)$ is inspired by Importance sampling. Training is based on maximising the discriminative objective
\beq
\log \tilde{p}(c_m|x_m,\theta) + \sum_{d\in\sett{N}}\log\br{1-\tilde{p}(d|x_m,\theta)}
\eeq
where $c_m$ is the correct class for input $x_m$ and $\sett{N}_c$ is a set of `negative' classes for $c_m$.  The objective is summed over all points in the minibatch. BlackOut shares similarities with NCE but avoids the difficulty of the unknown normalisation constant. For the IS distribution $Q(c)$ the authors propose to use $Q(c)\propto f(c)^\alpha$ where $f(c)$ is the empirically observed class distribution $f(c)\propto \sum_n \ind{c_n=c}$ and $0\leq \alpha<1$ is found by validation. BlackOut shares similarities with our normalisation approach. However, the training objective is different -- BlackOut uses a discriminative criterion rather than the likelihood. Whilst the optimum of the BlackOut objective can be shown to match the log likelihood objective (in the limit of a large number of samples) it is unclear why the BlackOut objective might be preferable to a direct log likelihood approximation. 
%As our experiments show, this is indeed the case. Whilst BlackOut can perform well, it does not outperform a simple log likelihood approximation approach.  

\begin{figure}[t]
\begin{center}
\subfigure[]{\includegraphics[width=0.45\tw]{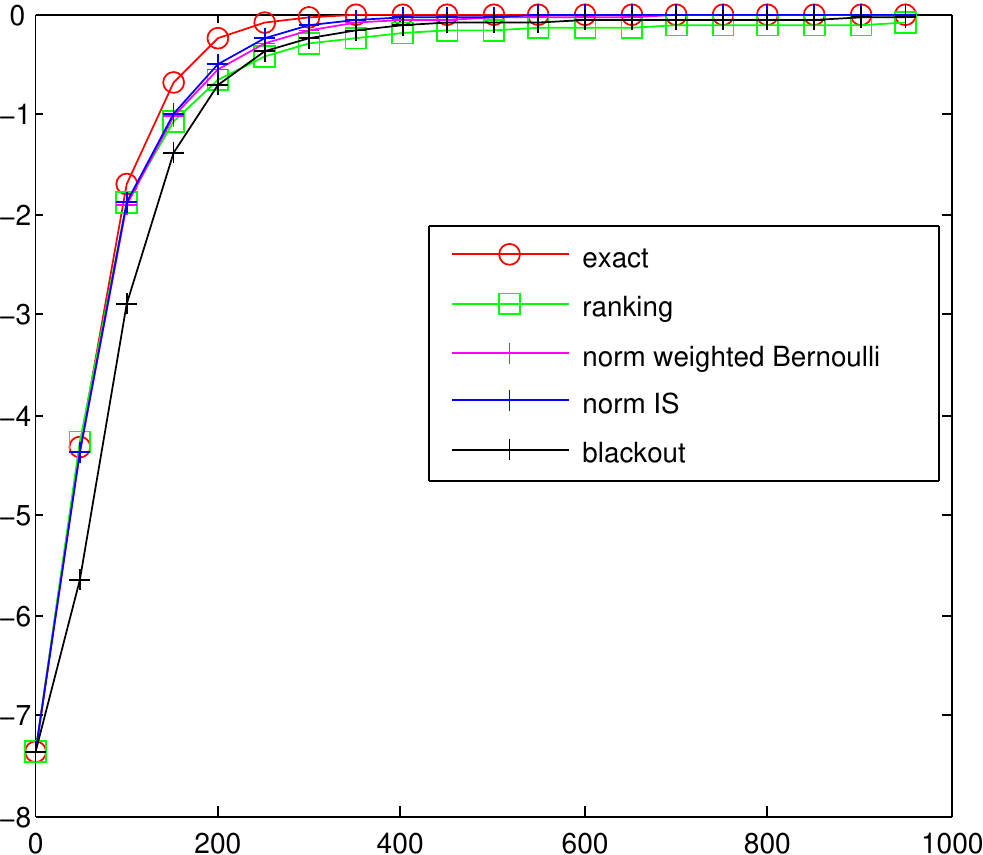}}
\subfigure[]{\includegraphics[width=0.45\tw]{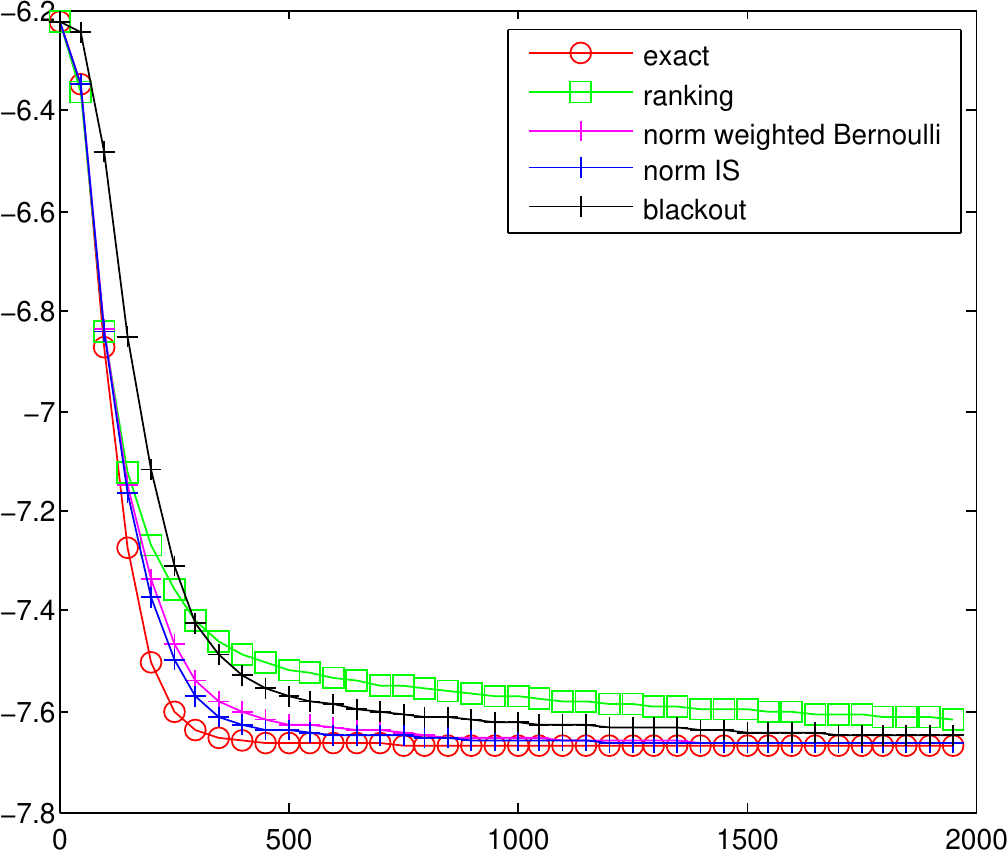}}
\end{center}
\caption{\label{fig:experiment} (a) Plotted is the value of the exact log likelihood (y-axis) against iteration number (x-axis) for a set of $N=2000$ datapoints, each $D=100$ dimensional. There are $C=1000$ classes and the training data was generated from the model to make a realisable problem. We compute the exact gradient for each minibatch of $|\sett{M}|=50$ datapoints and also compute the comparative and normalisation approximations for the minibatch. In all approximations, $K=20$ additional `negative' classes were randomly sampled in addition to the classes in each minibatch. All approximations used roughly $1050$ calculations of the form $\exp(\v{w}\trans\v{x})$ per minibatch, compared to 50,000 calculations for the exact approach, leading to a roughly 50 fold decrease in computation cost. The plot shows gradient ascent with momentum $0.99$.  Learning rates for the exact, and normalisation approximations were all set the same; the BlackOut and Ranking learning rates were set to the largest values that ensured convergence. The Noise Contrastive Estimation and Negative Sampling approaches are not shown since here $z_m=1$ results in very poor performance. (b) The log mean absolute difference between the current parameters learned based on the exact gradient update and the parameters learned by each approximation. This indicates the bias in learning the parameters. All approaches were initialised with the same initial parameters, so that at the first iteration there is no difference between the parameters.}
\end{figure}

\section{Experiment\label{sec:experiments}\label{sec:softmax:experiments}}

%\subsection{Softmax Regression\label{sec:softmax:experiments}}

We consider the simple softmax regression model $u(c,x)=\exp(\v{w}_c\trans\v{x})$. The exact log-likelihood in this case is concave, as are our sampling approximations.  This is useful since the convexity of the objective means that the results do not depend on the difficulty of optimisation and focus on the quality of the objective in terms of mimicking the true log likelihood.  To estimate the performance of the trained models, we note that the model is invariant up to $\v{w}_c\trans\v{x} + a(\v{x})$ for any $a(\v{x})$. This invariance means that directly comparing parameters from trained models is problematic. For this reason, we estimate bias by looking at the predictive performance of a model $p_\theta(c|x)$ compared to the predictive performance of the true model $p_{\theta_0}(c|x)$. In particular we measure the log mean absolute difference in class probability prediction for the inputs in the training set.  Neither Noise Contrastive Estimation (with $z_m=1$) nor Negative Sampling are given in the results since these approaches perform significantly worse than the other approaches.\newl

In \figref{fig:experiment} we show results for a simple experiment that compares the exact minibatch gradient compared to our normalisation approximations, ranking and BlackOut.  This experiment shows that whilst all methods work reasonably well, the normalisation approximations result in the most rapidly convergent in terms of bias minimisation. The empirical class frequency $Q(c)=f(c)$ was used to form the Importance sampling distribution and $K=20$ results in $b(c)=f(c)^{0.54}$ for the Bernoulli probabilities. There appears little difference between the Bernoulli and Importance sampling approaches which is somewhat unexpected. It is possible that the theoretical benefit of the Bernoulli sampling in terms of a lower variance estimator is dwarfed by the stochasticity induced by the minibatch sampling process.

\section{Discussion}

In contrast to recently introduced alternative approaches, a simple approximation of the the standard maximum likelihood objective provides an easily implementable and competitive method for fast large-class classification. \newl

An insight from our normalisation approximation is that it relates to the ranking objective and indeed justifies why an offset term can significantly improve the ranking objective. \newl

We also experimented with a deterministic approximations based on  based on variations of the result  
\begin{align}
Z&=\sum_{i=1}^C \exp{v_\theta(i,x)}=\sum_{i=1}^C \exp\br{\ave{\frac{s_iv_\theta(i,x)}{b_i}}{s_i}}\leq C+\ave{\sum_{i=1}^C {s_i\br{\exp\br{\frac{v_\theta(i,x)}{b_i}}-1}}}{\v{s}}
\end{align}
where $s_i$ are Bernoulli random variables and $b_i=p(s_i=1)$. However, these approaches were less successful in this context than the simple sampling approximations.

%\printbibliography

\bibliography{references} 
\bibliographystyle{plain}

\end{document}